\documentclass[11pt]{article}
 
\usepackage{arxiv}

\usepackage[utf8]{inputenc} 
\usepackage[T1]{fontenc}    
\usepackage{hyperref}       
\usepackage{url}            
\usepackage{booktabs}       
\usepackage{amsfonts}       
\usepackage{nicefrac}       
\usepackage{microtype}      
\usepackage{lipsum}		
\usepackage{graphicx}
\usepackage{natbib}
\usepackage{doi}

\usepackage{multirow}
\usepackage{algorithm}
\usepackage{algpseudocode}
\usepackage{graphicx}
\usepackage{xcolor}
\usepackage{soul}
\usepackage{amsmath}
\usepackage{amsfonts}
\usepackage{caption,subcaption}
\usepackage[para]{footmisc}
\usepackage{booktabs}
\usepackage[noadjust]{cite}
\usepackage{xcolor,colortbl}
\usepackage{hyperref}
\hypersetup{colorlinks = true, citecolor = blue }
\usepackage{cleveref}
\usepackage{caption}
\newcommand\blfootnote[1]{%
  \begingroup
  \renewcommand\thefootnote{}\footnote{#1}%
  \addtocounter{footnote}{-1}%
  \endgroup
}

\title{Textual Entailment Recognition with Semantic Features from Empirical Text Representation}

\author{ Md Shajalal$^{1,6}$, Md Atabuzzaman$^{2,4}$, Maksuda Bilkis Baby$^{2}$,\\ Md Rezaul Karim$^{1,3}$, Alexander Boden$^{1,5}$\\
    $^{1}$Fraunhofer Institute for Applied Information Technology FIT, Germany\\
	$^{2}$Hajee Mohammad Danesh Science and Technology University, Bangladesh\\
	$^{3}$RWTH Aachen University, Germany\\
	$^{4}$Bangladesh University of Engineering and Technology\\
	$^{5}$Bonn-Rhein-Sieg University of Applied Sciences, Germany\\
	$^{6}$University of Siegen, Germany\\
	\textit{atabuzzaman@gmail.com} \\
}

\renewcommand{\shorttitle}{Textual Entailment Recognition with Semantic Features}

\hypersetup{
pdftitle={Textual Entailment Recognition with Semantic Features from Empirical Text Representation},
pdfauthor={David S.~Hippocampus, Elias D.~Striatum},
}

\begin{document}
\maketitle

\begin{abstract}
Textual entailment recognition is one of the basic natural language understanding~(NLU) tasks. Understanding the meaning of sentences is a prerequisite before applying any natural language processing~(NLP) techniques to automatically recognize the textual entailment. A text entails a hypothesis if and only if the true value of the hypothesis follows the text. Classical approaches generally utilize the feature value of each word from word embedding to represent the sentences. In this paper, we propose a novel approach to identifying the textual entailment relationship between text and hypothesis, thereby introducing a new semantic feature focusing on empirical threshold-based semantic text representation. We employ an element-wise Manhattan distance vector-based feature that can identify the semantic entailment relationship between the text-hypothesis pair. We carried out several experiments on a benchmark entailment classification~(SICK-RTE) dataset. We train several machine learning~(ML) algorithms applying both semantic and lexical features to classify the text-hypothesis pair as entailment, neutral, or contradiction. Our empirical sentence representation technique enriches the semantic information of the texts and hypotheses found to be more efficient than the classical ones. In the end, our approach significantly outperforms known methods in understanding the meaning of the sentences for the textual entailment classification task.\blfootnote{This is the pre-print version of our accepted and presented paper at International Conference on Speech \& Language Technology for Low-resource Languages~(SPELLL’2022)}
\end{abstract}

\keywords{Textual entailment \and Semantic representation \and Word embedding \and Machine learning}

\section{Introduction}
Recognizing Textual Entailment~(RTE) is one of the basics of Natural Language Understanding~(NLU) and NLU is a subclass of Natural Language Processing~(NLP). Textual entailment is the relationship between two texts where one text fragment, referred to as \textit{`Hypothesis~(H)'} can be inferred from another text fragment, referred to as \textit{`Text (T)'}~\citep{dagan2005pascal,sharma2015recognizing}. In other words, Text $T$ entails Hypothesis $H$, if hypothesis $H$ is considered to be true according to the corresponding text $T$'s context~\citep{dagan2005pascal}. Let’s consider a text-hypothesis pair to illustrate an example of an entailment relationship. Suppose \textit{``A mother is feeding milk to a baby"} a particular text $T$ and \textit{``A baby is drinking milk"} is a hypothesis $H$. We see that the hypothesis $H$ is a true statement that can easily be inferred from the corresponding text $T$. Let's consider another hypothesis $H$, \textit{``A man is eating rice''}. For the same text fragment $T$, we can see that there is no entailment relationship between $T$ and $H$. Hence this text-hypothesis pair does not hold any entailment relationship, meaning neutral. The identification of entailment relationship has a significant impact in different NLP applications that include question answering, text summarization, machine translation, information extraction, information retrieval etc.~\citep{almarwani2017arabic,sharma2015recognizing}.

Since the first PASCAL challenge~\citep{dagan2005pascal} for recognizing textual entailment to date, different machine learning approaches have been proposed by the research community. The proposed approaches tried to employ supervised machine learning (ML) techniques using different underlying lexical, syntactic, and semantic features of the text-hypothesis pair. Recently, deep learning-based approaches including LSTM (Long Short Term Memory), CNN (Convolutional Neural Network), and Transfer Learning are being applied to detect the entailment relationship between the text-hypothesis pair~\citep{kiros2015skip,vaswani2017attention,devlin2018bert,conneau2017supervised}. Almost all methods utilized the semantic information of the text-hypothesis pair by representing them as semantic vectors. For doing so, they considered all the values of the words' vectors returned from the word embedding model. Classical approaches also apply the average of real-valued words' vectors as sentence representation. We hypothesize that some values of a particular vector of a word might impact negatively since they will be passed through an arithmetic average function. Considering this intuition, we observed that the elements of the words' vectors whose relevant elements are already present in the semantic vectors of the text-hypothesis pair, can be eliminated to get a better semantic representation. Following this observation, we proposed a threshold-based representation technique considering the mean and standard deviation of the words' vectors. 

Applying the threshold-based semantic sentence representation, the text and hypothesis are represented by two real-valued high-dimensional vectors. Then we introduce an element-wise Manhattan distance vector (EMDV) between vectors for text and hypothesis to have semantic representation for the text-hypothesis pair. This EMDV vector is directly employed as a feature vector to ML algorithms to identify the entailment relationship of the text-hypothesis pair. In addition, we introduce another feature by calculating the absolute average of the element-wise Manhattan distance vector of the text-hypothesis pair. In turn, we extract several handcrafted lexical and semantic features including Bag-of-Words (BoW) based similarity score, the Jaccard similarity score (JAC), and the BERT-based semantic textual similarity score (STS) for the corresponding text-hypothesis pair. To classify the text-hypothesis pair, we apply multiple machine learning classifiers that use different textual features including our introduced ones. Then the ensemble of the ML algorithms with the majority voting technique is employed that provides the final entailment relationship for the corresponding text-hypothesis pair. To validate the performance of our method, a wide range of experiments are carried out on a benchmark SICK-RTE dataset. The experimental results on the benchmark textual entailment classification dataset achieved efficient performance to recognize different textual entailment relations. The results also demonstrated that our approach outperforms some state-of-the-art methods. 

The rest of the paper is organized as follows:~\Cref{related work} presents some related works on RTE. Then our method is discussed in~\Cref{pa}. The details of the experiments with their results are presented in~\Cref{experiment and result}. Finally,~\Cref{conclusion} presents the conclusion with the future direction.

\section{Related Work} \label{related work}
With the first PASCAL challenge, textual entailment recognition has gained considerable attention of the research community~\citep{dagan2005pascal}. Several research groups participated in this challenge. But most of the methods applied lexical features~(i.e., word-overlapping) with ML algorithms to recognize entailment relation~\citep{dagan2005pascal}. Several RTE challenges have been organized and some methods with promising performance on different downstream tasks are proposed~\citep{haim2006second,giampiccolo2007third,giampiccolo2008fourth,bentivogli2009fifth,bentivogli2011seventh,dzikovska2013semeval,paramasivam2021survey}. Malakasiotis et al.~\citep{malakasiotis2007learning} proposed a method employing the string matching-based lexical and shallow syntactic features with support vector machine~(SVM). Four distance-based features with SVM are also employed~\citep{castillo2008approach}. The features include edit distance, distance in WordNet, and longest common substring between texts.

Similarly, Pakray et al.~\citep{pakray2009lexical} applied multiple lexical features including WordNet-based unigram match, bigram match, longest common sub-sequence, skip-gram, stemming, and named entity matching. Finally, they applied SVM classifiers with introducing lexical and syntactic similarity. Basak et al.~\citep{basak2015recognizing} visualized the text and hypothesis leveraging directed networks (dependency graphs), with nodes denoting words or phrases and edges denoting connections between nodes. The entailment relationship is then identified by matching the graphs' with vertex and edge substitution. Some other methods made use of bag-of-words, word-overlapping, logic-based reasoning, lexical entailment, ML-based methods, and graph matching to recognize textual entailment\citep{ghuge2014survey,renjit2022feature,liu2016classification}.

Bowman et al.~\citep{bowman2015large} introduced a Stanford Natural Language Inference corpus (SNLI) dataset consists of labeled sentence pairs that can be used as a benchmark in NLP tasks. This is a very large entailment (inference) dataset that provides the opportunity for researchers to apply deep learning-based approaches to identify the entailment relation between text and hypothesis. Therefore, different deep learning-based approaches including LSTM (Long Short Term Memory), CNN (Convolutional Neural Network), BERT, and Transfer Learning are being applied to RTE~\citep{kiros2015skip,vaswani2017attention,devlin2018bert,conneau2017supervised}. All the methods either used lexical or semantic features. But our proposed method uses both the lexical and semantic features including element-wise Manhattan distance vector~(EMDV), an average of EMDV, BoW, Jaccard similarity, and semantic textual similarity to recognize entailment.

\begin{figure*}[!ht]
\centering
    \includegraphics[width=0.95\textwidth]{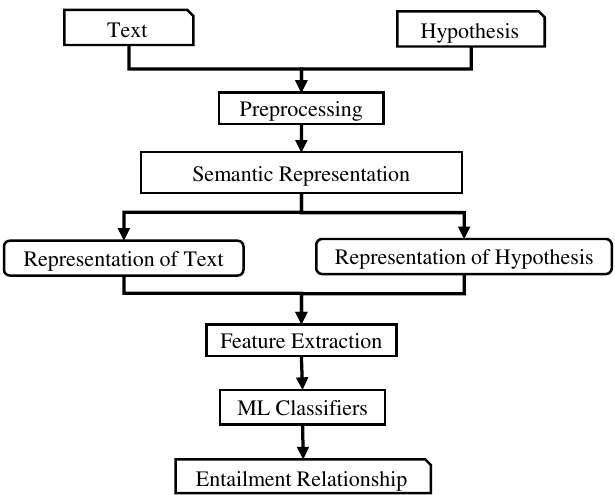}
    \caption{Overview diagram for recognising textual entailment} \label{overview}
\end{figure*} 
\section{Proposed Approach} \label{pa}
This section describes the proposed framework to recognize textual entailment (RTE) using the semantic information of the Text-Hypothesis (T-H) pair. The overview of our method is presented in ~\Cref{overview}. First, we apply different preprocessing techniques to have a better textual representation that eventually boosts the performance. In this phase, punctuation marks are removed, and the stopwords are also eliminated (except for negative words such as no, not, etc.). Here, a stopword is a word that has very little influence on the meaning of the sentence (i.g. a, an, the, and, or, etc.). After that, a tokenizer is utilized to split the sentence into a list of words. Then, we apply a lemmatizer to get the base form of the words.

\subsection{Empirical text representation}
The semantic information of a word is represented as a vector. The elements of the vector are real numbers that represent the contextual meaning of that word. By using the word-embedding, the semantic vectors of the words can be obtained. Almost all classical approaches apply arithmetic average using the words' semantic vectors to get the semantic information of the sentences. But all the values of the words' semantic vector might not be important to express the meaning of the text-hypothesis pair in the form of vectors. We hypothesize that some values of a particular vector of a word might impact negatively since they will be passed through an arithmetic average function.

Let $T$ and $H$ be the input text and hypothesis, respectively. We apply our sentence representation~\Cref{algo} to represent the sentence that provides better semantic information. From the first two statements, the function named $preprocess(T)$ returns the list of preprocessed words $T_p$ and $H_p$ for corresponding text $T$ and hypothesis $H$, respectively. After that, a vector of size $K$ with the initial value of zeros is taken (statement 3). Then a function ($get_-Semantic_-Info()$) returns the semantic representation applying our threshold-based empirical representation (statement 4-21).

\begin{algorithm*}[h!]
\caption{Semantic information of T-H pair based on an automated threshold of words semantic representation:} \label{algo} 
\begin{algorithmic}[1]
    \Require{Text (T) and Hypothesis (H) pair and Word-embedding models (word2vec)} 
    \Ensure{Semantic information of Text (T) and Hypothesis (H)}
    \State $T_p \Leftarrow preprocess(T)$ \Comment{List of words of Text}
    \State $H_p \Leftarrow preprocess(H)$ \Comment{List of words of Hypothesis} 
    \State $vec_-S \Leftarrow [0,0,...,0]$
    \Function{$get_-Semantic_-Info$}{$Sent$}
    \For{each $word \in Sent$}
        \If {$word \in word2vec.vocab$}
            \State $x \Leftarrow word2vec[word]$ \Comment{Vector representation of the word}
            \State $\overline{x} \Leftarrow Mean(x) $
            \State $\sigma \Leftarrow Standard\_Deviation (x) $
            \State $\alpha \Leftarrow \overline{x}+\sigma $
            \State $k \Leftarrow 0$
            \While {$k < length(x)$}
                \If { abs($vec_-S[k]-x[k]) \geq \alpha$}
                    \State $vec_-S[k] \Leftarrow add(vec_-S[k],x[k])$
                \EndIf
                \State $k++$
            \EndWhile
       \EndIf
    \EndFor
    \State $return$ $vec_-S$

    \EndFunction
\State $v_-T \Leftarrow get_-Semantic_-Info(T_p)$ \Comment{Semantic information of Text}
\State $v_-H \Leftarrow get_-Semantic_-Info(H_p)$ \Comment{Semantic information of Hypothesis}
\end{algorithmic}
\end{algorithm*}

In the function, the words which are available in the word2vec model vocabulary are considered for further actions. The first-word representation is added with $vec_-S$ without considering any condition or automated threshold. We found with empirical experiments that some elements from the word's vector might give a negative bias in the arithmetic average. By ignoring those, the sentence can be represented as a semantic vector omitting sampling fluctuation. Therefore, we attempt to study the sentence intent information by employing an automated threshold ($\alpha$) on the semantic elements of the words. We hypothesize that, if any particular element in index $i$ has not a significant absolute difference with the average of feature value in the same index $i$ in the sentence $S$, then the representation might not capture correct contextual information. In turn, we introduce the empirical threshold $\alpha$ using mean and standard deviation as $\alpha = \overline{x} + \sigma$. The elements of the word are added after employing the threshold to get the semantic information of the sentence. 

\subsection{Feature extraction of text-hypothesis Pair} 
\subsubsection{Element-wise Manhattan distance vector~(EMDV)}
The empirical threshold-based text representation returns the semantic real-valued vectors $v_T$ and $v_H$ for text and hypothesis, respectively. Our primary intuition to recognize the entailment relationship is that, the smaller the difference between text and hypothesis the larger the chance of entailment between them. Therefore we apply the Manhattan distance function to compute the element-wise Manhattan distance vector $EMDV = v_T - v_H$ where each element is the difference between the corresponding elements of the vectors for $T$ and $H$, respectively.

\subsubsection{Average of EMDV} The EMDV provides a real-valued Manhattan distance vector for the text-hypothesis pair. Applying the average over the summation of the absolute difference between text $v_T$ and hypothesis $v_H$ representations, we can calculate the average of EMDV which is a scaler value corresponding to the text-hypothesis pair. This can be calculated as following:

\begin{equation}\label{avgemdv}
    Sum_{EMDV} = \frac{1}{k}\sum_{i}^{k}{abs(v_{T_i} - v_{H_i})},
\end{equation}
where $k$ is the dimension of the vector. $v_{T_i}$ and $v_{H_i}$ are the $i$-th elements of the text and hypothesis, respectively.  

\subsubsection{Jaccard similarity score (JAC)} Jaccard similarity assesses the similarity of the text-hypothesis pair (T-H) to determine which words are common and which are unique. It is calculated by no. of common words present in the pair divided by no. of total words present in the sentence pair. This can be represented in set notation as the ratio of intersection ($T \cap H$) and union ($T \cup H$) of two sentences.

\begin{equation*}
    JAC(T,H) = \frac{T \cap H}{T \cup H}
\end{equation*}
where ($T \cap H$) indicates the number of words shared between both sentences and ($T \cup H$) provides the total number of words in both sentences (shared and un-shared). The Jaccard Similarity will be 0 if the two sentences don't share any values and 1 if the two sentences are identical.

\subsubsection{Bag-of-Words based similarity (BoW)} BoW is the vector representation where the dimension of the vector is the number of unique words exist in the text and the value of the vector is the frequency of the words. Suppose, two Bag-of-Words based vectors for text and hypothesis are obtained as $[1,0,2,0,4,0,0,0,1,1]$ and $[0,2,0,1,4,3,0,1,2,1]$ for T and H respectively. Then cosine similarity\citep{atabuzzaman2021semantic} is applied on these vectors to compute the similarity score.

\subsubsection{BERT-based semantic similarity score (STS)} Inspired by one of the prior works~\citep{shajalal2019semantic} on semantic textual similarity, we applied several semantic similarity methods. To compute the semantic textual similarity score (STS), pre-trained BERT word embedding is employed. Using the BERT word embedding, the T and H are represented as semantic vectors adding the words' vectors one by one. Then the cosine similarity between the vectors for text and hypothesis is considered as the STS score.

\section{Experiments Results} \label{experiment and result}
This section presents the details about the dataset, evaluation metrics, experimental setup, and performance analysis compared with known related works.

\subsection{Dataset}
We applied our method to a benchmark entailment recognition dataset named SICK-RTE~\citep{marelli2014sick}. This is an English dataset consisting of almost 10K English Text-Hypothesis (T-H) pairs exhibiting a variety of lexical, syntactic, and semantic phenomena. Each text-hypothesis pair is annotated as either \textbf{Neutral}, \textbf{Entailment} or \textbf{Contradiction} which are used as ground truth. Among 10k text-hypothesis pairs 5595 are annotated as Neutral, 2821 as Entailment, and 1424 as Contradiction. ~\Cref{dataset} presents some text-hypothesis pairs with corresponding entailment relations.

\begin{table*}[h!]
    \centering
    \caption{Examples of Text-Hypothesis pair from SICK-RTE dataset} \label{dataset} 
    \begin{tabular}{|p{4.5 cm}|p{4.5 cm}|p{2 cm}|} \hline 
    {\textbf{Text (T)}} &{\textbf{Hypothesis (H)}} &{\textbf{Relationship}}\\ \hline
    Two dogs are fighting.	 &{Two dogs are wrestling and hugging.} &{Neutral}\\ \hline
    A person in a black jacket is doing tricks on a motorbike. &{A man in a black jacket is doing tricks on a motorbike.} &{Entailment}\\ \hline
    Two dogs are wrestling and hugging.  &{There is no dog wrestling and hugging.} &{Contradiction}\\ \hline
    A woman selling bamboo sticks talking to two men on a loading dock. &{There are at least three people on a loading dock.} &{Entailment}\\ \hline
    A woman selling bamboo sticks talking to two men on a loading dock. &{A woman is selling bamboo sticks to help provide for her family.} &{Neutral}\\ \hline
    A woman selling bamboo sticks talking to two men on a loading dock. &{A woman is not taking money for any of her sticks} &{Contradiction}\\ \hline
    \end{tabular}
\end{table*}

We make use of the pre-trained BERT word-embedding model and pre-trained word-embedding model (word2vec) trained on the Google news corpus. The dimension of each word vector is $k=300$ and $k=768$ for word2vec and BERT, respectively. We evaluate the performance of our methods in terms of classification accuracy.

\subsection{Experimental settings} 
To evaluate the performance of our approach, several experiments have been carried out on the SICK-RTE dataset. First, we fed our element-wise Manhattan distance vector (EMDV)-based $k$ dimensional feature vector to the ML classifiers, and this setting is denoted as $RTE_-EMDV$. The setting where the sentence representation did not apply the threshold-based algorithm is denoted as ${RTE_{without_-thr}}_{-}EMDV$. Then we compute the feature vector for the text-hypothesis pair employing the Average of EMDV, JAC, BOW, and STS measures. To illustrate the performance of our element-wise Manhattan distance vector, we applied the Average of EMDV~(ref.~\Cref{avgemdv}) in two different variations with the threshold-based algorithm and with a plain vector from embedding. For all settings, we employed several classification algorithms including support vector machine with RBF kernel, K-nearest neighbors, random forest, and naive Bayes. Finally, the ensemble result considering the majority voting of the ML algorithms is also considered. To do the experiments 75\% data are used in training and the rest are used as testing data.

\subsection{Performance analysis of entailment recognition}
\Cref{entailmentmd} demonstrates the performance of different ML algorithms to recognize entailment relation with element-wise Manhattan distance vector-based features ($RTE_-EMDV$). Here we also reported the performance of the EMDV feature vector without applying the representation algorithm (~\Cref{algo}), (${RTE_{without_-thr}}_{-}EMDV$). This table illustrates that the KNN classifier achieved better performance than other ML algorithms to detect entailment relationship using the representational $RTE_-EMDV$. The table also demonstrates how the element-wise distance-based feature vector from threshold-based semantic representation helps the ML models to recognize different entailment labels.

For a better understanding of the impact of the our introduced features, we present the~\Cref{Conf1} which is the confusion  matrixes of the ensemble methods considering the element-wise EMDV vector with and without the proposed sentence representation algorithm. \Cref{cmwiththrmd} shows that the ensemble method can detect neutral, entailment, and contradiction T-H pair. But~\Cref{cmwithoutthrmd} reflects that without the proposed representation ML algorithms are not able to recognize the contradiction relationship between text and hypothesis and only can detect a few entailment relations. This also signifies the impact of our proposed feature with threshold-based sentence representation. 

\begin{table*}[h!]
    \centering
    \caption{Performance of ML models based on EMDV}
    \begin{tabular}{|c|c|c|} \hline
    \textbf{Algorithm} &\textbf{Features} &\textbf{Accuracy} \\ \hline
    \multirow{2}{*}{SVM\_rbf} &$RTE_-EMDV$ &0.66  \\
    &${RTE_{without_-thr}}_{-}EMDV$ &0.58  \\ \hline
    \multirow{2}{*}{KNN} &$RTE_-EMDV$ &\textbf{0.67}  \\
    &${RTE_{without_-thr}}_{-}EMDV$ &\textbf{0.57}  \\ \hline
    \multirow{2}{*}{R.Forest} &$RTE_-EMDV$ &0.62  \\
    &${RTE_{without_-thr}}_{-}EMDV$ &0.58  \\ \hline
    \multirow{2}{*}{Naive Bayes} &$RTE_-EMDV$ &0.66  \\
    &${RTE_{without_-thr}}_{-}EMDV$ &0.58  \\ \hline
    \multirow{2}{*}{Ensemble} &$RTE_-EMDV$ &0.66  \\
    &${RTE_{without_-thr}}_{-}EMDV$ &0.58  \\ \hline
    \end{tabular}
    
    \label{entailmentmd}
\end{table*}

\begin{table*}[!htb]
    \caption{Confusion matrix for ensemble learning with EMDV}
    \begin{minipage}{.5\linewidth}
      \caption{With threshold}
      \label{cmwiththrmd}
      \centering
       \begin{tabular}{|c|c|c|c|} \hline
        &Neutral &Entail &Contradict  \\ \hline
        Neutral &\textbf{1135}  &193   &17 \\ \hline
        Entail &328  &\textbf{370}   &37\\ \hline
        Contradict &96   &155  &\textbf{129}\\ \hline
    \end{tabular}
    \end{minipage}%
    \begin{minipage}{.5\linewidth}
      \centering
        \caption{Without threshold}
        \label{cmwithoutthrmd}
        \begin{tabular}{|c|c|c|c|} \hline
            &Neutral &Entail &Contradict  \\ \hline
            Neutral &\textbf{1407}  &15   &0 \\ \hline
            Entail &674  &\textbf{13}   &0\\ \hline
            Contradict &344   &7  &0\\ \hline
        \end{tabular}
    \end{minipage} 
    \label{Conf1}
\end{table*}

\if false
\begin{table}[h!]
    \centering
    \caption{Confusion Matrix of the Ensemble measure using proposed threshold based M\_D}
    \begin{tabular}{|c|c|c|c|} \hline
        &Neutral &Entail&Contradiction  \\ \hline
        Neutral &\textbf{1407}  &15   &0 \\ \hline
        Entailment &674  &\textbf{13}   &0\\ \hline
        Contradiction &344   &7  &0\\ \hline
    \end{tabular}
    \label{cmwiththrmd}
\end{table}

\begin{table}[h!]
    \centering
    \caption{Confusion Matrix of the Ensemble measure without proposed threshold based M\_D}
    \begin{tabular}{|c|c|c|c|} \hline
        &Neutral &Entailment &Contradiction  \\ \hline
        Neutral &\textbf{1407}  &15   &0 \\ \hline
        Entailment &674  &\textbf{13}   &0\\ \hline
        Contradiction &344   &7  &0\\ \hline
    \end{tabular}
    \label{cmwithoutthrmd}
\end{table} 
\fi
\Cref{rm} presents the performance of different ML algorithms to recognize entailment with semantic and lexical features including our proposed average of EMDV feature (\Cref{avgemdv}). The table reflects that when all the features are combined, all the classifiers are showing better performance with the average of EMDV feature (\Cref{avgemdv}) than without threshold-based text representation. This also consistently demonstrates that the proposed average of the EMDV feature can capture a better entailment relationship than other classical features. \Cref{conf2} presents the confusion matrices of the ensemble models with different features' combinations. Both tables show that with all the features considering the proposed semantic representation, the ensemble method can classify different text-hypothesis pairs more accurately than classical semantic representation. This also concludes the performance consistency.

\begin{table}[h!]
    \centering
    \caption{Performance of ML models based on handcrafted features}
    \begin{tabular}{|c|c|c|} \hline
    \textbf{Algorithm} &\textbf{Features} &\textbf{Accuracy} \\ \hline
    \multirow{2}{*}{SVM\_rbf} &${Avg_-Sum}_-{EMDV}$+BoW+JAC+STS &0.80  \\
    &${Avg_-Sum}_-{without_-thr}$+BoW+JAC+STS &0.78  \\ \hline
    \multirow{2}{*}{KNN} &${Avg_-Sum}_-{EMDV}$+BoW+JAC+STS &0.81  \\
    &${Avg_-Sum}_-{without_-thr}$+BoW+JAC+STS &0.79  \\ \hline
    \multirow{2}{*}{R.Forest} &${Avg_-Sum}_-{EMDV}$+BoW+JAC+STS &0.81  \\
    &${Avg_-Sum}_-{without_-thr}$+BoW+JAC+STS &0.78  \\ \hline
    \multirow{2}{*}{Naive Bayes} &${Avg_-Sum}_-{EMDV}$+BoW+JAC+STS &0.74  \\
    &${Avg_-Sum}_-{without_-thr}$+BoW+JAC+STS &0.73  \\ \hline
    \multirow{2}{*}{Ensemble} &${Avg_-Sum}_-{EMDV}$+BoW+JAC+STS &\textbf{0.81}  \\
    &${Avg_-Sum}_-{without_-thr}$+BoW+JAC+STS &\textbf{0.79}  \\ \hline
    \end{tabular}
    
    \label{rm}
\end{table}
\if false
\begin{table}[h!]
    \centering
    \caption{Confusion matrix for ensemble prediction}
    \begin{tabular}{|c|c|c|c|} \hline
        &Neutral &Entailment &Contradiction  \\ \hline
        Neutral &\textbf{1252}  &149   &25 \\ \hline
        Entailment &191  &\textbf{479}   &14\\ \hline
        Contradiction &124   &15  &\textbf{211}\\ \hline
    \end{tabular}
    \label{cmwithoutthr}
\end{table}

\begin{table}[h!]
    \centering
    \caption{Confusion matrix for ensemble prediction using all features and threshold-based semantic information}
    \begin{tabular}{|c|c|c|c|} \hline
        &Neutral & Entailment & Contradiction  \\ \hline
        Neutral & \textbf{1225}  & 138   & 16 \\ \hline
        Entailment & 177  &\textbf{523}   &8\\ \hline
        Contradiction & 108   & 23  &\textbf{242}\\ \hline
    \end{tabular}
    \label{cmwiththr}
\end{table}
\fi

\begin{table}[!htb]
    \caption{Confusion matrix for ensemble learning with handcrafted features}
    \begin{minipage}{.5\linewidth}
      \caption{With threshold}
      \centering
       \begin{tabular}{|c|c|c|c|} \hline
         &Neutral &Entail &Contradict   \\ \hline
            Neutral  &\textbf{1225}  &138   &16 \\ \hline
            Entail &177  &\textbf{523}   &8\\ \hline
            Contradict &108   &23  &\textbf{242}\\ \hline
    \end{tabular}
    \label{cmwithoutthr}   
    \end{minipage}%
    \begin{minipage}{.5\linewidth}
      \centering
        \caption{Without threshold}
        \begin{tabular}{|c|c|c|c|} \hline
            &Neutral &Entail &Contradict  \\ \hline
            Neutral &\textbf{1252}  &149   &25 \\ \hline
            Entail &191  &\textbf{479}   &14\\ \hline
            Contradict &124   &15  &\textbf{211}\\ \hline
           
        \end{tabular}
    \end{minipage}
    \label{conf2}
\end{table} 
\begin{figure*}[!ht]
    \centering
    \includegraphics[width=0.75\linewidth]{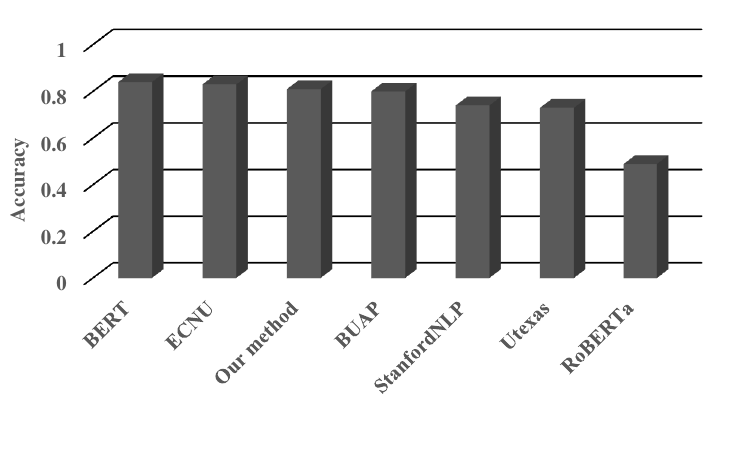}
    \caption{Performance comparison of our proposed method in terms of Accuracy on the SICK-RTE dataset.} \label{compare_entailment}
\end{figure*}

\subsection{Comparative analysis}
\Cref{compare_entailment} shows the comparison of different known prior related works with our method on the SICK-RTE dataset. BUAP~\citep{bentivogli2016sick} employed a language model with different features including sentence's syntactic and negation features to classify text-hypothesis pairs as entailment, neutral, and contradiction. Utexas~\citep{bentivogli2016sick} used sentence composition and phrase composition features with negation and vector semantic model to recognize the entailment relationship. These two models employed different features and neural network models but still, our method outperformed them. Our proposed features with an empirical threshold-based sentence representation algorithm can capture better semantic entailment relationships. Different feature engineering method with deep learning is employed for RTE task by ECNU~\citep{shin2020autoprompt}. BERT~(Finetuned)~\citep{shin2020autoprompt} applied bidirectional encoder representation of sentences pair. Though our method does not outperform ECNU and BERT(Finetuned), the performance difference compared to them is subtle and hence performed effectively. 

\section{Conclusion with future direction} \label{conclusion}
This paper presents a novel method to recognize textual entailment by introducing new features based on element-wise Manhattan distance vector employing empirical semantic sentence representation technique. To extract the semantic representation of the text-hypothesis pair, an empirical threshold-based algorithm is employed. The algorithm eliminates the unnecessary elements of the words' vectors and extracts the semantic information from the text-hypothesis pair. Then various ML algorithms are employed with the extracted semantic information along with several lexical and semantic features. The experimental results indicate the efficiency in identifying textual entailment relationships between text-hypothesis pairs. In summary, the performance of different experimental settings with multiple classifiers was consistent and outperformed some know-related works.

In the future, it would be interesting to apply deep learning-based method using the element-wise Manhattan distance vector to recognize text entailment.

\bibliographystyle{unsrtnat}
\bibliography{references}  






\end{document}